\documentclass[sigconf]{aamas}

\usepackage{latexml}

\usepackage{balance}
\usepackage{makecell}
\usepackage{caption}
\usepackage{graphicx}
\usepackage{amsmath}
\usepackage{tabularx}
\usepackage{array}
\usepackage{booktabs}
\usepackage{arydshln}
\usepackage{xcolor}
\AtBeginDocument{
  \setlength{\abovedisplayskip}{4pt plus 2pt minus 2pt}
  \setlength{\belowdisplayskip}{4pt plus 2pt minus 2pt}
  \setlength{\abovedisplayshortskip}{0pt plus 2pt}
  \setlength{\belowdisplayshortskip}{2pt plus 2pt minus 2pt}
}

\makeatletter
\renewcommand\footnotetextcopyrightpermission[1]{}
\makeatother

\makeatletter
\newcommand\blfootnote[1]{%
  \begingroup
  \renewcommand\thefootnote{}%
  \def\Hy@raisedlink##1{}%
  \footnotetext{#1}%
  \endgroup
}
\makeatother

\iflatexml
\else
\setcopyright{none}
\copyrightyear{}
\acmYear{}
\acmDOI{}
\acmISBN{}
\fi

\title{Automated Multi-Agent Organization Design: \texorpdfstring{\\}{} From Architecture Search to Organizational Learning}

\iflatexml
  \author{Shao-Jun Xia\hyperlink{equal}{\textsuperscript{*}}}
\else
  \author{Shao-Jun Xia}%
  \authornote{Equal contribution.}
\fi
\affiliation{
  \institution{Duke University}
  \city{Durham}
  \country{United States}
}
\email{shaojun.xia@duke.edu}

\iflatexml
  \author{Yuner Zhang\hyperlink{equal}{\textsuperscript{*}}}
\else
  \author{Yuner Zhang}
  \authornotemark[1]
\fi
\affiliation{
  \institution{Carnegie Mellon University}
  \city{Pittsburgh}
  \country{United States}}
\email{yunerz@andrew.cmu.edu}

\iflatexml
  \author{Huixin Zhang\hyperlink{corresponding}{\textsuperscript{\textdagger}}}
\else
  \author{Huixin Zhang}
  \authornote{Corresponding authors.}
\fi
\affiliation{
  \institution{Texas A\&M University}
  \city{College Station}
  \country{United States}}
\email{zhanghui21@tamu.edu}

\iflatexml
  \author{Di Wang\hyperlink{corresponding}{\textsuperscript{\textdagger}}}
\else
  \author{Di Wang}%
  \authornotemark[2]
\fi
\affiliation{
  \institution{Foxconn}
  \city{Milwaukee}
  \country{United States}
}
\email{di.wang@foxconn.com}

\author{Yifan Jiang}
\affiliation{
  \institution{University of Southern California}
  \city{Los Angeles}
  \country{United States}}
\email{yifjia@isi.edu}

\iflatexml
  \author{Liangxi Liu\hyperlink{corresponding}{\textsuperscript{\textdagger}}}
\else
  \author{Liangxi Liu}%
  \authornotemark[2]
\fi
\affiliation{
  \institution{Northeastern University}
  \city{Boston}
  \country{United States}
}
\email{liu.liangx@northeastern.edu}

\begin{abstract}\looseness=-1
Agentic system design is increasingly shifting from hand-crafted prompts and workflows toward automated search over agentic programs and interaction structures, with multi-agent systems receiving growing attention. Yet a central question remains unresolved: when does an automatically generated multi-agent structure exhibit a functionally meaningful division of responsibility? We argue that automated multi-agent design should move beyond searching for effective architectures toward identifying which responsibilities are functionally necessary, how they should be allocated, and when they should change. To explore this, we propose Automated Multi-Agent Organization Design (AMOD), which treats organization itself as a first-class design object, explicitly modeling how agent teams are composed, what functional responsibilities the agents assume, how they coordinate, and how collective decisions are made. We formulate AMOD as an execution-driven design loop. For each task, an automated designer instantiates an organization by assigning functional responsibilities and coordination rules to a set of agents. After execution, task outcomes, intermediate artifacts, and interaction traces are examined to identify which responsibilities may be missing, duplicated, or ineffective, and the organization is revised accordingly.
We also propose testing discovered responsibilities by removing them, replacing them with generic roles, or changing their assignment under matched resource budgets, with complementary evaluations. AMOD provides a principled foundation for evidence-based and adaptive multi-agent organization design.
\end{abstract}


\begin{document}

\fancypagestyle{standardpagestyle}{%
  \fancyhf{}%
  \fancyfoot[C]{\thepage}%
  \renewcommand{\headrulewidth}{0pt}%
  \renewcommand{\footrulewidth}{0pt}%
}
\fancypagestyle{firstpagestyle}{%
  \fancyhf{}%
  \fancyfoot[C]{\thepage}%
  \renewcommand{\headrulewidth}{0pt}%
  \renewcommand{\footrulewidth}{0pt}%
}
\pagestyle{standardpagestyle}
\maketitle

\iflatexml
  {\footnotesize
  \noindent\hypertarget{equal}{\textsuperscript{*}}Equal contribution.\\
  \hypertarget{corresponding}{\textsuperscript{\textdagger}}Corresponding authors.\par}
\fi

\section{Introduction}
\label{sec:intro}

\begin{figure}[t]
  \centering
  \includegraphics[width=\columnwidth]{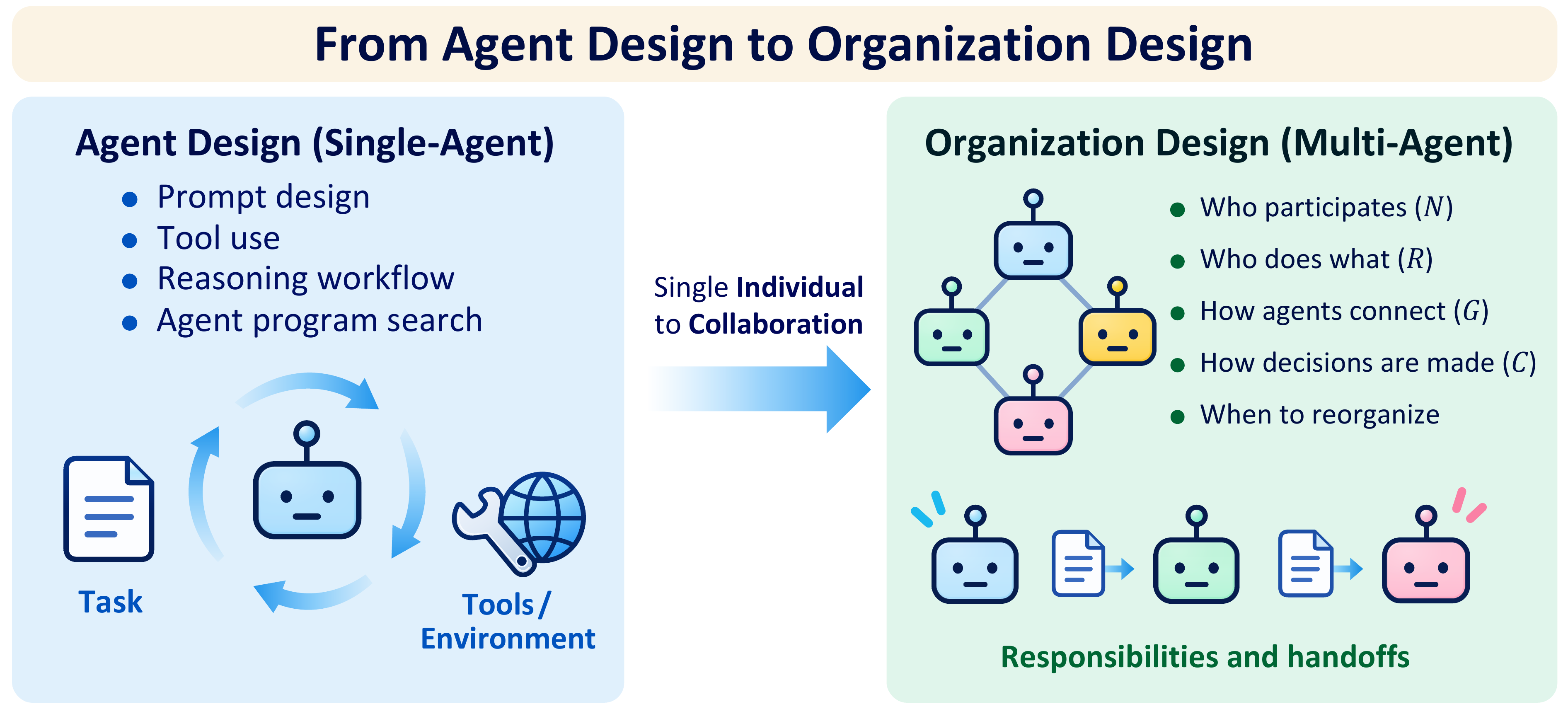}
  \caption{\textbf{From agent design to organization design.} AMOD jointly designs team composition, functional responsibilities, coordination, collective decisions, and adaptation.}
  \Description{Two panels show single-agent design through prompts, tools, workflows and program search, and multi-agent organization design through composition, responsibilities, coordination, decisions and reorganization. A blue gradient arrow connects how to compute to collaboration.}
  \label{fig:paradigm}
  \vspace{-8pt}
\end{figure}

Recent work has begun to automate design decisions specified by hand in agentic systems. Automated Design of Agentic Systems (ADAS) treats the agentic program as an object of search, allowing a meta-agent to generate and improve reasoning procedures, tool use, workflows, and their combinations~\cite{hu2025automated}. For multi-agent systems, the design problem also concerns how agents divide work and interact. Methods now search over computational graphs~\cite{zhuge2024language}, query-dependent architectures~\cite{zhang2025multi}, team composition and communication structures~\cite{zhou2026multi,li2026assemble}, and evolving roles and interactions~\cite{wang2026metagen}. These choices can improve task performance and resource use, but aggregate outcomes alone do not identify which organizational functions account for the gains.
Fig.~\ref{fig:paradigm} summarizes the shift from designing the computation of a single agent to organizing collaboration.

Separately, automated failure attribution~\cite{zhang2025agent}, trajectory-level credit assignment~\cite{wuinteraction}, and contribution attribution~\cite{jiang2026semantic} provide evidence about execution. Other work revises roles from traces~\cite{song2026abstral}, adapts roles and interactions during inference~\cite{wang2026metagen}, or compares organizational variables under controlled settings~\cite{chen2026toward}. These advances leave open which comparisons can establish the value of a responsibility suggested by execution evidence.
Addressing this gap requires an organizational view of multi-agent systems. A workflow specifies how computation and information flow, but may leave the obligation of each agent implicit. An organizational view represents participating agents together with their functional responsibilities, coordination mechanisms, and decision procedures, consistent with longstanding multi-agent organization research~\cite{horling2004survey,horling2008using}. Making responsibility explicit provides a unit that can be examined, reassigned, removed, or compared under controlled conditions.

We formulate \textbf{Automated Multi-Agent Organization Design (AMOD)} as the problem of automatically constructing, evaluating, and revising organizations around explicit functional responsibilities. Code-level search such as ADAS~\cite{hu2025automated} can already express organizational designs, but AMOD makes their functional obligations and validation conditions explicit. AMOD asks whether multiple agents are needed, how functions should be allocated and coordinated, and what should change when execution reveals missing, redundant, or ineffective functions. Evidence for \emph{functional necessity} comes from measurable degradation when a function is removed or altered, relative to explicitly specified, budget-matched alternatives. Such a claim is conditional on the tested task, agents, and alternatives.
AMOD considers how to form and maintain useful organizations as tasks, agent populations, and environments change. Execution, diagnosis, and revision address the current organization, while experience from earlier tasks may inform later designs. We use responsibility discovery and validation as a tractable entry point, asking whether functional divisions of labor add value beyond what additional agents, tokens, or samples provide.

\section{What Constitutes a Multi-Agent Organization?}
\label{sec:organization}

\begin{figure}[t]
  \centering
  \includegraphics[width=\columnwidth]{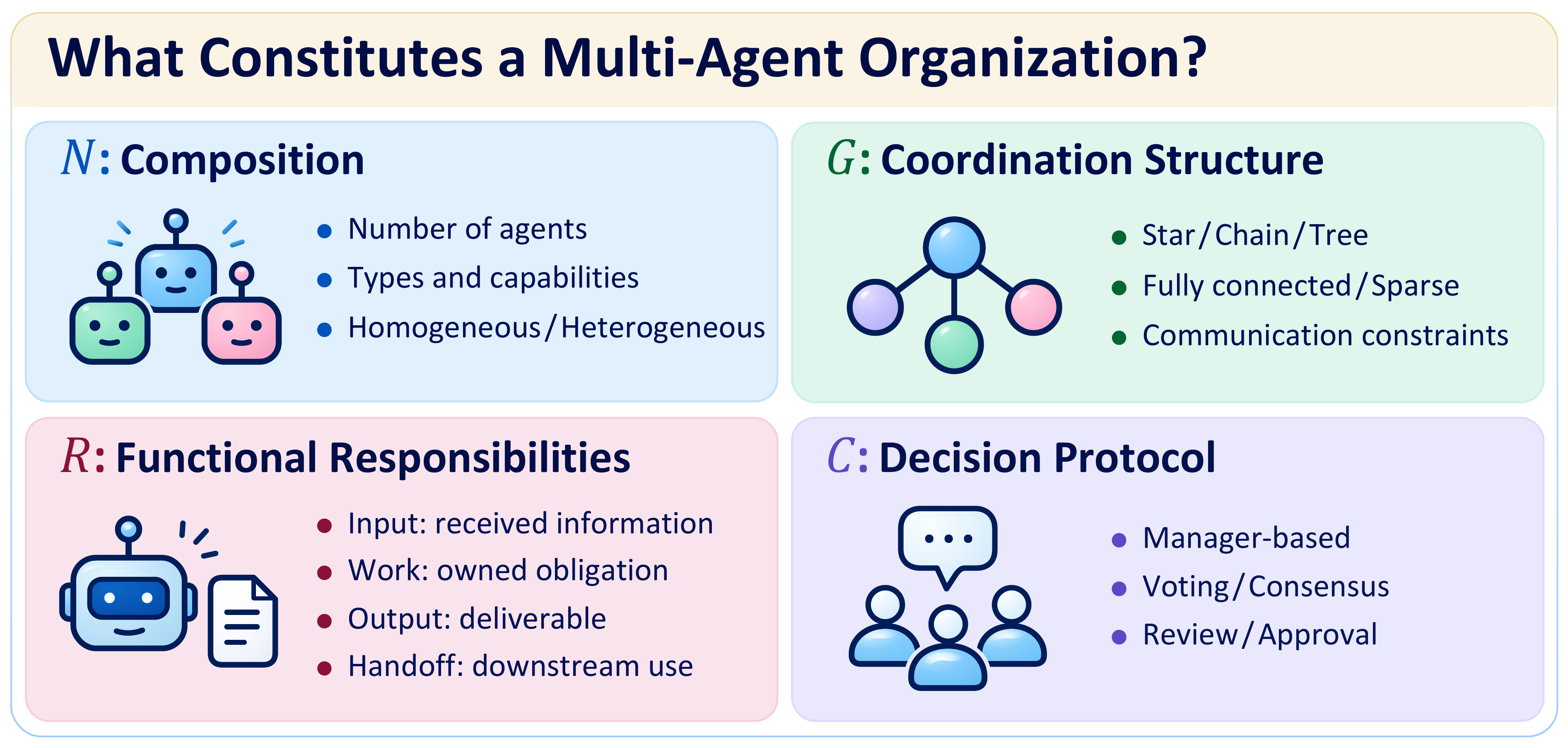}
  \caption{\textbf{Four organizational dimensions.} Responsibilities define executable functions. Identical communication structures can support different responsibilities and decision protocols.}
  \Description{Four pastel cards: N composition (number, types, homogeneity of agents); G coordination structure (star, chain, tree, fully connected, sparse graph); R functional responsibilities (input, obligation, output, handoff); C decision protocol (manager-based, voting or consensus, review and approval).}
  \label{fig:organization}
  \vspace{-6pt}
\end{figure}

Since AMOD treats the organization itself as a design object, organizational design choices must be represented explicitly. As illustrated in Fig.~\ref{fig:organization}, we represent an organization as
\begin{equation}
D=(N, G, R, C),
\end{equation}
where $N$ is the set of participating agent instances, with team size $n=|N|$ and a composition specified by their capabilities; $G$ is the coordination structure through which agents communicate; $R$ records functional responsibilities and their agent assignments; and $C$ specifies how outputs are combined, conflicts resolved, and collective decisions made~\cite{horling2004survey,horling2008using}.
The dimensions are not independent, as $N$ constrains feasible structures $G$ and assignments $R$. In practice, $G$ may be a star, chain, or tree~\cite{zhu2025multiagentbench}, a fully connected debate~\cite{du2024debate}, or a sparse graph~\cite{zhang2025agentprune}. And $C$ turns partial results into a collective outcome through manager decisions~\cite{wu2024autogen,zhu2025multiagentbench}, voting or consensus~\cite{du2024debate}, or review and approval~\cite{kim2026teambench}. The same coordination structure can support different responsibilities and decision protocols, and thus different organizations (Fig.~\ref{fig:organization}).

The responsibility component $R$ is central here. It pairs each responsibility specification $r$ with an assigned agent $a\in N$. The specification records inputs, obligation, deliverable, recipient, and completion condition, beyond a label such as Planner or Coder. In a software task, for example, an interface-coordination responsibility returns compatibility checks or repair requirements before component integration. Its function, the work it is meant to contribute to the organization, can be checked, reassigned, or replaced independently of its executor. Task ownership and handoff obligations belong to $R$, whereas $G$ supplies the interaction channels.
For a fixed available population, a designer $\pi_{\theta}$ with parameters $\theta$ proposes $D^{*}=\pi_{\theta}(x)$ for task $x$, including both the participating agents and their organization. We allow $n=1$, so that the designer must determine whether collaboration is needed.
This compact core leaves fine-grained information access, authority, organizational memory, and runtime adaptation outside the immediate representation. We focus first on discovering a useful division of functional responsibility, and Sec.~\ref{sec:outlook} considers how the design problem changes when these further choices are included.

\section{Discovering and Validating Functional Responsibilities}
\label{sec:loop}

\begin{figure*}[t]
  \centering
  \includegraphics[width=0.961\textwidth]{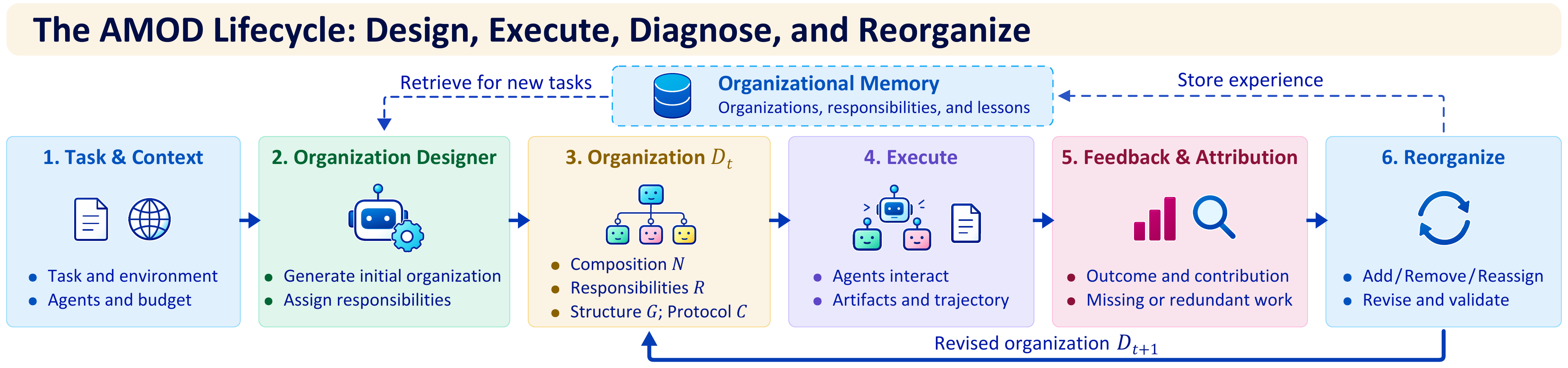}
  \caption{\textbf{The AMOD lifecycle.} Task context guides design, while execution and attribution motivate reorganization. Solid arrows track task-level revision, while dashed arrows show future retention and retrieval of experience across tasks.}
  \Description{Six stages connect task and context, organization designer, organization, execution, feedback and attribution, and reorganization. Reorganization feeds the next design iteration. Organizational memory retains and retrieves experience across tasks.}
  \label{fig:loop}
  \vspace{-6pt}
\end{figure*}

Generating an organization does not establish the value of its responsibilities, as a new responsibility may fill a coordination gap or merely duplicate existing work. Comparisons with strong single-agent and repeated-sampling baselines must control resources to distinguish organizational benefits from gains due to additional computation~\cite{jwalapuram2026illusion,xu2026rethinking,dylan2026equal}. Responsibility-level comparisons share execution budget caps and access to information and tools, with calls, tokens, samples, and latency reported. Latency is an outcome, not a matching constraint. Comparisons of complete designers additionally count proposal, diagnosis, revision, and validation calls used to select responsibilities. Budget released by removing a responsibility can be reallocated within the organization or used for repeated sampling.

We hold models and tools fixed within each comparison, although failures may still reflect agent capability, as discussed in Sec.~\ref{sec:outlook}. Given task $x$ and organization $D_t=(N_t, G_t, R_t, C_t)$ at design iteration $t$, execution yields a trajectory $\tau_t=\operatorname{Execute}(x, D_t)$ of outputs, messages, tool calls, artifacts, and environmental feedback (Fig.~\ref{fig:loop}). Contrasting traces can suggest specialist roles~\cite{song2026abstral}, and contribution attribution can identify agents supporting valuable outcomes~\cite{jiang2026semantic}. These observations motivate hypotheses about missing functions. Failed components, downstream artifact use, duplicated work, and blocked handoffs provide evidence for a revision:
\begin{equation}
D_{t+1}=\operatorname{Reorganize}(D_t, \tau_t, z_t),
\end{equation}
where $z_t$ denotes diagnostic evidence used to justify the change, such as failed observable checks, coordination problems, or estimated contributions.

Consider two agents whose local implementations pass observable checks but whose merged solution fails an observable integration check because of incompatible interface assumptions. The designer proposes an interface-coordination responsibility $r^*$ that compares the interfaces and returns repair requirements before integration. Its inputs are the components and their interface requirements, and its deliverable identifies incompatible assumptions for the integrating agents. Completion means checking each shared interface and stating any required repairs, rather than producing another feature. Assigning it to an existing agent $a^*\in N_t$ gives $R_{t+1}=R_t\cup\{(r^*, a^*)\}$ without increasing team size.

Validation compares $D_{t+1}$ with removal, generic-role replacement, and reassignment variants, rerunning all conditions from the same initial task state with the same available feedback. Removal tests the value of the function, generic-role replacement tests whether the explicit responsibility specification adds value, and reassignment tests executor suitability. Trajectory analysis must distinguish a removed function from one taken over by another agent. Similarly, if a generic reviewer performs the same checks, it supplies a substitute implementation rather than showing that checking is unnecessary. A loss after reassignment may reflect a mismatch between the agent and the responsibility. These tests support retaining $r^*$ only within the conditions and alternatives examined. They may also support merging responsibilities or using no explicit coordination.
Responsibility-level tests keep $N$, $G$, and $C$ fixed where possible. If removing a function requires changing the coordination structure or a decision rule, the result concerns a joint intervention and cannot be attributed to $R$ alone.

\section{A Tractable Research Program}
\label{sec:program}

\begin{table*}[t]
\centering
\footnotesize
\setlength{\tabcolsep}{4pt}
\renewcommand{\arraystretch}{1.05}
\caption{\textbf{Complementary evaluation environments for AMOD.} Some proposed comparisons require adapted benchmark runners.}
\label{tab:testbeds}
\begin{tabular}{@{}llll@{}}
\toprule
\textbf{Testbed} & \textbf{What is primarily fixed} & \textbf{What can vary} & \textbf{Primary relevance to AMOD} \\
\midrule
MultiAgentBench~\cite{zhu2025multiagentbench} & Task setting and evaluator & Coordination topology and interaction strategy & Controlled diagnosis and organizational revision \\
CooperBench~\cite{khatua2026cooperbench} & Repository tasks, tools, and tests & Team composition and functional responsibilities & Responsibility discovery and validation \\
ChinaTravel~\cite{shao2026chinatravel} & Task interface and evaluator & Task-specific responsibilities and coordination & Cross-domain transfer of the design procedure \\
\noalign{\vskip 1.5pt}
\hdashline
\noalign{\vskip 1.5pt}
SILO-BENCH~\cite{zhang2026silo} & Task generator and per-agent shard size & Agent scale and communication protocol & Information distribution and scalability under silos \\
TeamBench~\cite{kim2026teambench} & Role boundaries and task environment & Presence or removal of functional roles & Responsibility isolation and role-level ablation \\
CoCoBench~\cite{chen2026cocobench} & Task instances and number of agents & Coordination mechanisms & Allocation, ordering, mutual exclusion, and handoff \\
\bottomrule
\end{tabular}
\end{table*}

The proposed evaluation uses benchmarks as task, tool, and evaluator interfaces, and AMOD chooses $D=(N, G, R, C)$ on top of them. It separately assesses task performance, responsibility-level effects, and design quality with models and tools fixed. The first three environments in Table~\ref{tab:testbeds} form the main program, while the others probe information distribution, role isolation, and coordination.

\textbf{Controlled diagnosis and revision.} MARBLE, the evaluation framework accompanying MultiAgentBench~\cite{zhu2025multiagentbench}, supplies task assignment, execution traces, and coordination structures. Here AMOD varies $D$ while keeping task rules, tool and information access, and models fixed, and compares otherwise matched designers that receive trace-informed or score-only feedback, to test whether richer feedback improves organizational revision.

\textbf{Responsibility discovery and validation.} CooperBench~\cite{khatua2026cooperbench} supplies repository-grounded feature pairs whose implementations may interact or conflict. We propose a joint-task adaptation in which the designer receives the requirements of both features and chooses responsibilities and integration procedures. Every method submits one final patch against the same hidden feature tests, and the benchmark runner performs no repair. All baselines receive identical requirements and tool access and use this evaluator. This differs from the published cooperative setting, in which agents implement separate features and their patches are merged before testing, so the published scores are not directly comparable. The responsibility-level interventions then examine the functions in the resulting organization.
Development and evaluation are split by repository. Designer parameters and revision rules are frozen before held-out evaluation, although within-task changes to $D_t$ remain allowed using observable feedback. Prior experience is not retrieved, and hidden tests are used only for final scoring.

\textbf{Cross-domain design procedure.} ChinaTravel~\cite{shao2026chinatravel} differs from CooperBench in inputs, tools, and output format. With the representation, designer parameters, revision procedure, and budget fixed, AMOD must generate planning responsibilities rather than reuse coding roles. The benchmark adapter only exposes tools and formats itineraries, without planning or repair. The test is whether the procedure discovers useful divisions, such as parallel transport and lodging planning with shared time and budget checks, validated through removal and replacement.

\textbf{Baselines and measures.} Comparisons include a strong single agent, a single agent given the same responsibility specifications and workflow~\cite{xu2026rethinking}, budget-matched repeated sampling~\cite{wang2023selfconsistency,brown2024monkeys}, fixed coordination structures, manually designed heterogeneous teams, and automated multi-agent system designers. These comparisons ask whether trajectory evidence improves design over scores alone, whether explicit responsibilities outperform generic alternatives, and whether the complete designer outperforms a strong single agent after accounting for design costs. Task scores are complemented by measures of duplicated work, handoff success, and artifact use. Inference cost, latency, communication volume, and search overhead are also reported. When the designer repeatedly selects $n=1$, budget-matched comparisons with multi-agent baselines assess whether collaboration would have improved outcomes on those tasks.
Repeated executions on the same instances give paired differences and uncertainty estimates. Trace-informed designers share the same observation interfaces and revision budget. A feedback ablation holds the designer and budget fixed while restricting feedback to observable scalar scores. Search stops when the budget is exhausted or no tested candidate improves the task metric, using development-set metrics during development and only observable checks on held-out tasks. Rejected proposals and validation used for selection count toward design cost, while repetitions used only for final statistical analysis are reported separately.

\section{Related Work}
\label{sec:related}

\textbf{Organizations.} Classical work formalizes task allocation~\cite{smith1980contractnet}, searches organizational structures~\cite{horling2004survey,horling2008using,sims2008automated}, and specifies roles, norms, and permissions~\cite{hubner2002moise,noriega2004engineering}. Foundation-model frameworks implement prescribed roles and conversations~\cite{wu2024autogen,hong2024metagpt,qian2024chatdev}. AMOD builds on this view while asking how execution evidence can motivate new functions and how interventions can test their value.

\textbf{Optimization targets.} The survey by Mad\v{z}ar and Mekterovi\'c~\cite{madzar2026adas_survey} distinguishes prompt, compound-system, workflow, modular, and code-level search. APE~\cite{zhou2023ape} generates and selects instructions, OPRO~\cite{yang2024opro} uses scored candidates, ProTeGi~\cite{pryzant2023protegi} uses textual gradients, and EvoPrompt~\cite{guo2024evoprompt} uses evolutionary operators. DSPy~\cite{khattab2024dspy} and MIPRO~\cite{opsahlong2024mipro} optimize compound language model programs, while TextGrad~\cite{yuksekgonul2025textgrad} and Trace~\cite{cheng2024trace} use richer feedback across computation steps. Other methods search workflows and graphs~\cite{zhuge2024language,zhang2025aflow,wang2025scoreflow}, modular architectures~\cite{shang2025agentsquare,zhang2025multi,saadfalcon2025archon}, agent code~\cite{hu2025automated,yin2025godelagent,zhang2025darwingodel}, populations~\cite{yuan2025evoagent}, and topologies~\cite{zhou2026multi,li2026assemble}. These targets describe what can change but do not by themselves establish the functional necessity of the resulting responsibilities.

\textbf{Evidence and adaptation.} Attribution methods localize failures and estimate contributions~\cite{zhang2025agent,wuinteraction,jiang2026semantic}. Lu et al.~\cite{lu2026agents} show that agent contribution estimates depend on the intervention protocol, coalition distribution, and evaluation objective. ABSTRAL~\cite{song2026abstral} derives specialists from contrasting traces, and MetaGen~\cite{wang2026metagen} adapts roles and interactions during inference. MANTA~\cite{huang2026manta} audits collaboration traces, applies bounded structural repairs, and accumulates experience across runs. Organizational studies separate design factors~\cite{chen2026toward}, and cost-aware evaluation motivates strong resource-matched baselines~\cite{kapoor2025agentsmatter,jwalapuram2026illusion,xu2026rethinking,dylan2026equal}. Agent memory architectures~\cite{park2023generativeagents} and multi-agent memory perspectives~\cite{yu2026memoryarchitecture} inform our discussion of organizational memory in Sec.~\ref{sec:outlook}. AMOD uses attribution to propose functions, budget-matched interventions to assess the conditional value of specified responsibilities, and validated cases as candidates for organizational memory.

\section{Toward Adaptive and Cumulative Organizations}
\label{sec:outlook}

The program above studies a tractable part of AMOD, namely organization design for a particular task and agent population. A broader account must address accumulated experience, changing capabilities, and organizational constraints beyond communication structure. A configuration may cease to be useful as these conditions change. The design problem therefore includes deciding which responsibilities to retain, revise, or discard, and which lessons from earlier executions should inform later tasks.

\textbf{Organizational memory.} A designer could retain useful organizations and revision evidence, building on prior work on reusable design knowledge~\cite{huang2026manta,song2026abstral}. A retained case should record task conditions, responsibility specifications, intervention outcomes, and budget, so that the designer can check whether the conditions supporting its value hold in the new setting. A transferable lesson can specify which assumptions require joint checking, such as interface compatibility in software or shared time and budget constraints in travel planning. This is more general than a domain-specific role name. Conflicting or outdated cases require revision, and a failed trajectory alone should not become a general design rule. On held-out CooperBench repositories, cold-start and retrieval-assisted design can be compared by success rate, revision count, and cost, using memory built from development repositories and frozen before evaluation. Retrieved responsibilities are revalidated under changed conditions. Memory is enabled in this separate reuse experiment, not in the primary evaluation of Sec.~\ref{sec:program}.

\textbf{Changing agent populations.} A team with a strong verifier and weak implementers may need a different allocation from a team of strong specialists, and roles can change during inference~\cite{wang2026metagen}. A designer may thus condition on task $x$, population $P$, environment $E$, and experience $H$, giving $D^{*}=\pi_{\theta}(x, P, E, H)$. Generalization can be tested by replacing one member while keeping tasks and tools fixed, and comparing the original allocation, reassignment, and redesign. If a responsibility remains useful after reassignment, its value is less tied to one particular executor, whereas persistent failure of the replacement agent across allocations would suggest a capability limitation rather than a missing organizational function. The tests can also vary communication cost or available resources.

\textbf{Authority and institutional constraints.} Agents acting on tools and shared state may modify resources, approve results, delegate tasks, or block actions of other agents. Electronic institutions model such roles, norms, and permissions~\cite{noriega2004engineering}. Design must then decide who should perform a function, who may perform it, and who must verify or approve the result. For example, an interface checker may request a repair without authority to edit shared code, so approval and implementation remain separate responsibilities.

\textbf{When organization is unnecessary.} Coordination can cost more than specialization saves for weakly coupled subtasks. Deciding this is hard, since failures may reflect weak models, successes may hide redundant agents, and incorrect diagnoses may contaminate memory. Knowing when not to organize matters as much as organizing better.

Across these extensions, the reusable object is a tested relation between task conditions, a function, and its allocation. Its value must be checked again when the agent population or operating constraints change.

\section{Conclusion}

In this paper, we have introduced AMOD, a framework that treats functional responsibilities as explicit and testable components of multi-agent organization. AMOD uses execution evidence to propose organizational revisions and evaluates them through budget-matched removal, generic-role replacement, and reassignment tests. These interventions help distinguish gains attributable to a particular responsibility from those produced by additional agents or computation, while revealing when a simpler or single-agent solution is sufficient. By retaining the conditions and evidence associated with validated responsibilities, the framework can further support organizational learning as tasks, agent capabilities, and resource constraints change. AMOD thus moves automated multi-agent design beyond architecture search toward evidence-based divisions of labor.

\clearpage
\bibliographystyle{unsrtnat}
\bibliography{references}

\end{document}